\documentclass[letterpaper, 10 pt, conference]{ieeeconf}  % Comment this line out if you need a4paper

\IEEEoverridecommandlockouts                              % This command is only needed if 
\usepackage{epsfig}
\usepackage{graphicx, picinpar,wrapfig}
\usepackage{amsmath}
\usepackage{amssymb}
\graphicspath{{../image/}}
\usepackage{ctable}
\usepackage{multirow, subfigure}
\usepackage{capt-of}
\usepackage{ctable}
\DeclareGraphicsExtensions{.pdf,.jpeg,.png,.eps,.jpg,.jpeg}
\usepackage{multirow}
\usepackage{hhline,xcolor}
\usepackage{makecell}
\DeclareGraphicsExtensions{.pdf,.jpeg,.png,.eps,.jpg,.jpeg}
\DeclareMathOperator*{\argmaxA}{arg\,max} % Jan Hlavacek
\DeclareMathOperator*{\argminA}{arg\,min}
\usepackage{caption}
\title{\LARGE \bf
	Multi-Sensor 3D Object Box Refinement for Autonomous Driving}

\author{Peiliang Li, Siqi Liu and Shaojie Shen% <-this % stops a space
	\thanks{All authors are with the Department of Electronic and Computer Engineering, The Hong Kong University of Science and Technology, Hong Kong, SAR China. {\tt\small \{pliap, sliubq, eeshaojie\}@ust.hk}}
}

\begin{document}

	\maketitle
	\thispagestyle{empty} %forces first page to not have a header
	\pagestyle{empty}
	
	%\ieeefootline{Workshop on Latex Style Files \\ International Conference on Latex 2014, Las Vegas, NV, USA}%creates footline
	
	%\ieeeheadline{Workshop on Latex Style Files \\ International Conference on Latex 2014, Las Vegas, NV, USA}%creates headline

	%%%%%%%%%%%%%%%%%%%%%%%%%%%%%%%%%%%%%%%%%%%%%%%%%%%%%%%%%%%%%%%%%%%%%%%%%%%%%%%%
	\begin{abstract}
		%Using this general geometric representation, the 3D object can be locally reconstructed with dense pixels or sparse 3D points. Applying the reconstructed shape to the raw sensor data, the 3D object location can be unified optimized via minimizing the stereo photometric error or point cloud alignment error.
		We propose a 3D object detection system with multi-sensor refinement in the context of autonomous driving. In our framework, the monocular camera serves as the fundamental sensor for 2D object proposal and initial 3D bounding box prediction. While the stereo cameras and LiDAR are treated as adaptive plug-in sensors to refine the 3D box localization performance. For each observed element in the raw measurement domain (e.g., pixels for stereo, 3D points for LiDAR), we model the local geometry as an instance vector representation, which indicates the 3D coordinate of each element respecting to the object frame. Using this unified geometric representation, the 3D object location can be unified refined by the stereo photometric alignment or point cloud alignment. We demonstrate superior 3D detection and localization performance compared to state-of-the-art monocular, stereo methods and competitive performance compared with the baseline LiDAR method on the KITTI object benchmark.
		
	\end{abstract}
	
	% %%%%%%%%%%%%%%%%%%%%%%%%%%%%%%%%%%%%%%%%%%%%%%%%%%%%%%%%%%%%%%%%%%%%%%%%%%%%%%%%
	\section{INTRODUCTION}
	3D object detection and localization have drawn increasing attention in these years as being a core perceptual function for self-driving vehicles. 
	%Existing 3D object detection works usually explore the use of specific input such as LiDAR point cloud \cite{chen2017multi,qi2017frustum,zhou2017voxelnet,ku2017joint,liang2018deep,Liang2019CVPR, wang2019frustum}, stereo imagery \cite{3dopJournal,li2018semantic, li2019stereo, wang2018pseudo}, or monocular image \cite{chen2016monocular,mousavian20173d,zeeshan2014cars,chabot2017deep,cvpr18xu, qin2018monogrnet, ku2019monocular}.
	%From the robotics perspective, however, it is not reliable to use a single sensor as the essential perception module for real-world applications. The monocular camera is cheap and informative but inaccurate in 3D localization. Stereo cameras are relatively low cost and provide effective depth information, but have a limited sensing range (depends on the focal length and the baseline length). LiDAR has accurate 3D localization ability, but less informative and sensitive to reflection (e.g., rainy, car window). 
	Plenty of recent efforts have been made to detect 3D objects using the monocular camera \cite{chen2016monocular,mousavian20173d,zeeshan2014cars,chabot2017deep,cvpr18xu, qin2018monogrnet, ku2019monocular}, as it is low-cost and provides rich semantic information for scene understanding. However, a single camera is naturally inaccurate in 3D localization. There are also other works exploring the use of specific depth sensor such as stereo imagery \cite{3dopJournal,li2018semantic, li2019stereo, wang2018pseudo}, which are also relatively low-cost and provide effective depth information, but have a limited sensing range; and LiDAR \cite{chen2017multi,qi2017frustum,zhou2017voxelnet,ku2017joint,liang2018deep,Liang2019CVPR, wang2019frustum}, which has accurate 3D localization ability, but is less informative and sensitive to reflection (e.g., rainy, car window).
	To achieve robust perception, modern self-driving vehicles tend to equip multiple different sensors, where the 3D information is represented in quite different ways (e.g., high-level semantic cues for the monocular image, pixel-level disparity for stereo images, sparse but geometric-aware point cloud for LiDARs).
	However, there is currently no explicit formulation to model the geometric relations between the object state and different kinds of raw sensor observations.
	As a result, naturally incorporating multiple sensors with variant characteristics brings researchers new challenges for designing a robust 3D object estimator.
	%As a result, dealing with multiple sensors brings researchers new challenges: how to develop a unified and flexible 3D object detection method that supports different kinds of sensors and even for their combinations? Furthermore, how to naturally fuse different sensor data to exploit their complementary potentiality? 
	Although the 3D bounding box can be separately regressed in end-to-end manners, employing individual networks for each sensor leads to redundant computation, and the missing of original uncertainty information poses barriers for the future sensor fusion.
	In another aspect, directly learning the multi-sensor model by stacking all the sensor data into a deep neural network (DNN) might lack the interpretability and the flexibility.
	
	\begin{figure}
		\begin{center}
			\includegraphics[width=0.93\columnwidth]{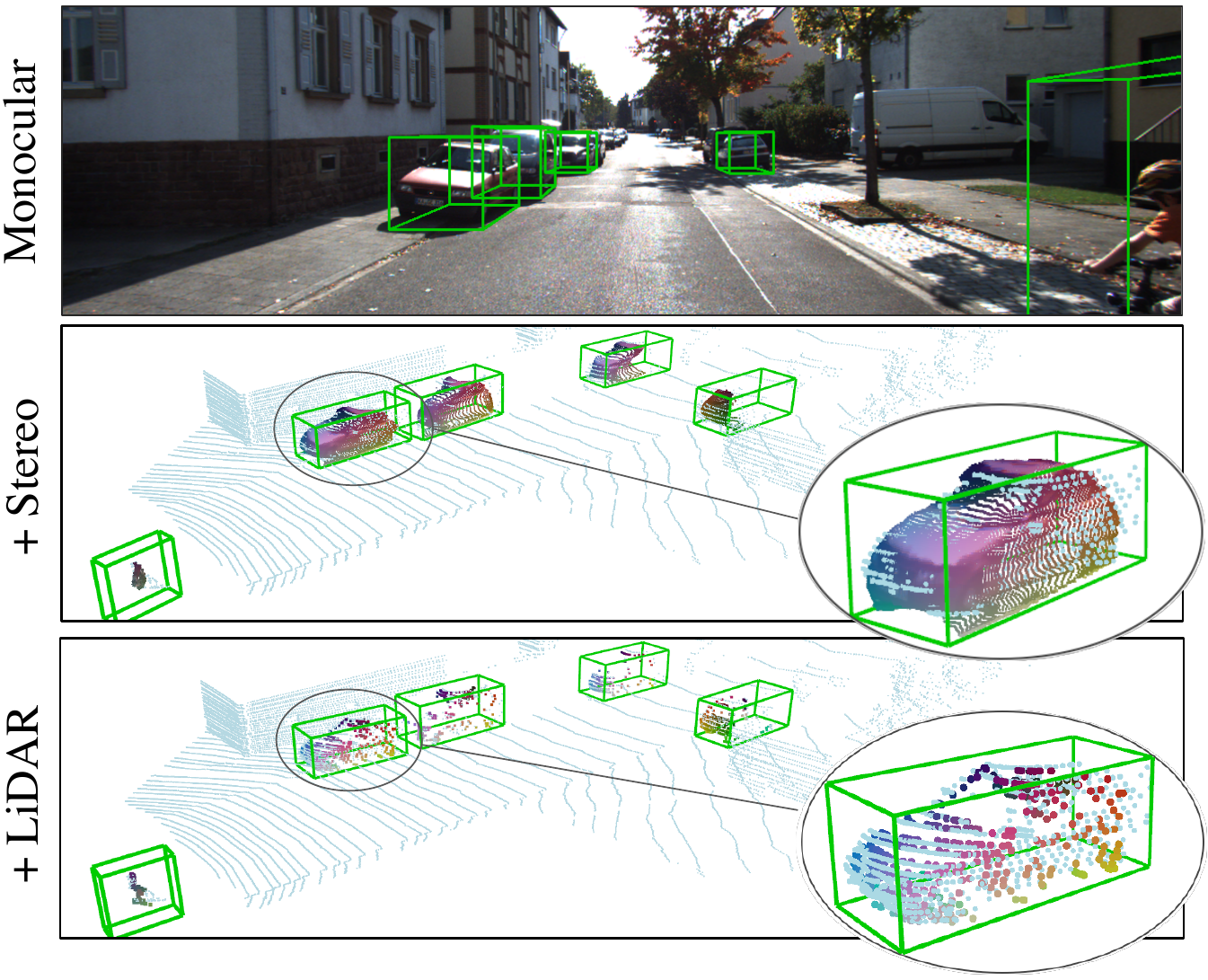}
		\end{center}
		\setlength{\belowcaptionskip}{-0.4cm}
		\caption{Example 3D object estimation results of our framework. From top to bottom: Our fundamental monocular 3D detection; Stereo refinement using the photometric alignment; LiDAR refinement using the point cloud alignment. Enlarged views show the pixel-wise or point-wise object shape which is used for geometric refinement.}
		\label{fig:cover}
		
	\end{figure}

	\begin{figure*}
		\begin{center}
			\vspace{0.2cm}
			\includegraphics[width=1.9\columnwidth]{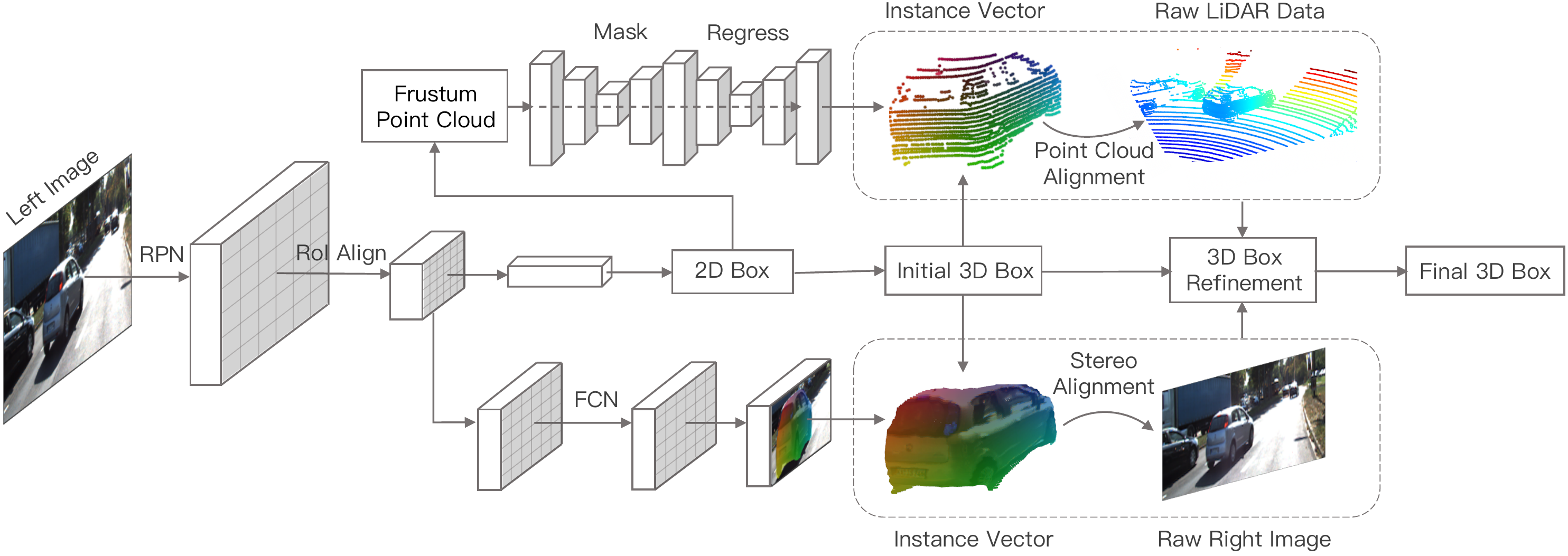}
		\end{center}
		\caption{System architecture of our 3D object detection and refinement, where we design a light-weight monocular 3D detector (Sect.~\ref{sec:mono}) to provide 2D RoIs and initial 3D boxes. Recovering the object shape using our instance vector representation, we perform unified 3D box refinement using the stereo photometric alignment (Sect.~\ref{sec:stereo}), point cloud alignment (Sect.~\ref{sec:lidar}).}
		\label{fig:system}
	\end{figure*}
	
	To this end, instead of focusing on the single modal accuracy, we aim to explicitly model the 3D geometry characterization in different sensor domain, where each sensor should be exploited their natural properties to contribute to the final 3D object estimation. Following this motivation, we design our two-stage system as shown in Fig.~\ref{fig:system}. Given a single image as the fundamental input, we extend the Faster R-CNN \cite{ren2015faster} to a light-weight monocular 3D object detector, where we predict several spatial information together with the 2D box and the object class to infer the initial 3D bounding box (Sect. \ref{sec:mono}). To unified utilize stereo or LiDAR to refine the 3D object estimation, we model the object structural information in the image or the point cloud into a unified instance vector representation, which indicates the 3D coordinate of each raw sensor element (e.g., pixels, 3D points) respecting to the object local frame. Specifically, if stereo images are available, we employ a region-based FCN to predict the object instance vector and segmentation mask after \textit{RoIAlign} \cite{he2017mask} layer. Anchoring the masked instance vector to the initial 3D bounding box, we obtain dense object-level 3D reconstruction (bottom half of Fig.~\ref{fig:system}). The initial 3D box accuracy can then be improved by warping the shape-aware object patch from the left image to the right image and optimizing the minimum photometric error. If the point cloud is given, we adopt PointNet \cite{qi2017pointnet} to predict the point-wise instance vector. Similar with \cite{qi2017frustum}, we take the point cloud contained in the object frustum as input and perform the mask segmentation and instance vector regression sequentially. With the point-wise instance vector, each 3D point is associated with the object center to form a sparse local shape under the same representation (top half of Fig.~\ref{fig:system}). Similarly, the 3D object can be further relocalized by reprojecting the local structure to the raw LiDAR point cloud and minimizing the point-wise distance.
	
	The instance vector has low-variance and spatial-invariant characteristics under the same object category. It is deterministically associated with the local appearance (e.g., the left-rear lamp of vehicles always appears at roughly same locations in vehicle frame) and thus considered as a suitable task for neural networks. Thanks to this unified representation, stereo cameras and the LiDAR can be treated as adaptive \textit{plug-in} sensors to improve the 3D object estimation using similar energy-minimization approaches. Based on the raw sensor uncertainty, we are further able to naturally fuse multiple raw sensor measurements to jointly refine the 3D object location. 
	
	Our key contributions are summarized as 1) A simple and effective monocular 3D object detector which serves as our fundamental and backup perceptual function. 2) The first to propose a unified geometric way to utilizing the stereo images or point cloud to refine the 3D object box. 3) Demonstrate state-of-the-art 3D detection and localization performance for both monocular, stereo, and show competitive 3D performence using the LiDAR on the KITTI object benchmark.
	
	\section{Related Work}
	%We briefly review recent works of 3D object detection based on the LiDAR data, monocular image and stereo images respectively.
	
	\subsection{3D Object Detection from Images.} Several works focus on using a single image to estimate 3D objects. They either utilize the context information such as shape prior, ground plane, and segmentation \cite{chen2016monocular}, or exploit geometry correspondences between 2D and 3D bounding box\cite{mousavian20173d}, or fit the sparse keypoint to wireframe models \cite{chabot2017deep, zeeshan2014cars}. 
	Recently, \cite{cvpr18xu,qin2018monogrnet} propose end-to-end multi-level fusion strategies for 3D object detection, where \cite{cvpr18xu} takes advantages of multidimensional input fusion, and \cite{qin2018monogrnet} uses multi-stage feature map fusion. To complement the missed depth information in monocular images, some works exploit the stereo imagery to achieve better localization accuracy.
	\cite{3dopJournal, wang2018pseudo} take the stereo-generate depth image as input, where \cite{3dopJournal} focuses on 3D proposals generation and scoring by encoding prior and context information, and \cite{wang2018pseudo} processes depth into pseudo point cloud, then employs existing LiDAR-based methods \cite{qi2017frustum, ku2017joint} to detect 3D objects.
	In another aspect, \cite{li2018semantic,li2019stereo} exploit the depth information from raw stereo images. \cite{li2018semantic} formulates spatial and temporal sparse correspondences into an object-level bundle adjustment problem. \cite{li2019stereo} takes advantage of both sparse and dense constraints in raw stereo images to achieve 3D object localization.

	\subsection{3D Object Detection from LiDAR.} Majority of existing 3D object detection methods rely on LiDAR to obtain accurate 3D information while representing LiDAR data in various ways.
	\cite{engelcke2017vote3deep,li20173d,luo2018fast,zhou2017voxelnet,yan2018second} quantize the raw point cloud into structured voxel grid, then use either 2D or 3D CNN to detect the 3D object.  \cite{li2016vehicle,luo2018fast, yang2018pixor} encode the point cloud into 2D image formats by projecting raw 3D points to front view or bird's eye view images, the 3D object detection can thereby be achieved using the regular 2D convolutional network. Benefited from the PointNet \cite{qi2017pointnet}, \cite{qi2017frustum, shi2018pointrcnn} propose to directly localize 3D objects in raw point cloud based on 2D proposals \cite{qi2017frustum} or 3D proposals \cite{shi2018pointrcnn}.

	\subsection{Using Multimodal Data.} There are also some works exploiting multiple sensor fusion for 3D object detection. \cite{chen2017multi, ku2017joint} extract feature for RGB image and LiDAR-project images in separate branches then fuse the feature map by element-wise mean in the following R-CNN stage, while \cite{liang2018deep} continuously fuses the multiple sensor streams during the feature extraction stage. Instead of converting LiDAR into 2D representations, \cite{xu2018pointfusion} processes RGB image and point cloud using ResNet \cite{he2016deep} and PointNet \cite{qi2017pointnet} respectively, then employs a fusion layer to perform dense fusion. However, all the above methods realize the multimodal fusion in the network which takes specified data as input, thus are not flexible enough, i.e., one trained model weight can only be applied to a fixed sensor combination.
	
	\section{3D Object Estimation Framework}
	
	In this section, we start with our monocular 3D object detector, then describe how to extend it to the stereo and point cloud refinement.
	A 3D bounding box is represented with its center location $\mathbf{p}_o$ = $[x_o, y_o, z_o]$, dimension $\mathbf{d}$ = $[w, h, l]$, and rotation $\mathbf{R}(\theta)$ parameterized by the horizontal orientation $\theta$. As illustrated in Fig.~\ref{fig:system}, our 3D object detection framework uses the monocular image to produce 2D RoI (Region of Interest) proposals and initial 3D bounding boxes, which is described in Sect.~\ref{sec:mono}. If additional sensor inputs (e.g., stereo images, point cloud) are available, we predict element-wise mask and instance vector for each proposal, 
	%the 3D object estimation can be unified formulated into a learning-aided energy-minimization problem, 
	the 3D object location can then be unified refined by minimizing the stereo photometric error (Sect.~\ref{sec:stereo}) or point cloud alignment error (Sect.~\ref{sec:lidar}).
	
	\subsection{Monocular 3D Object Detection}
	\label{sec:mono}
\begin{figure}
		\centering
		\vspace{0.2cm}
		\includegraphics[width=0.95\linewidth]{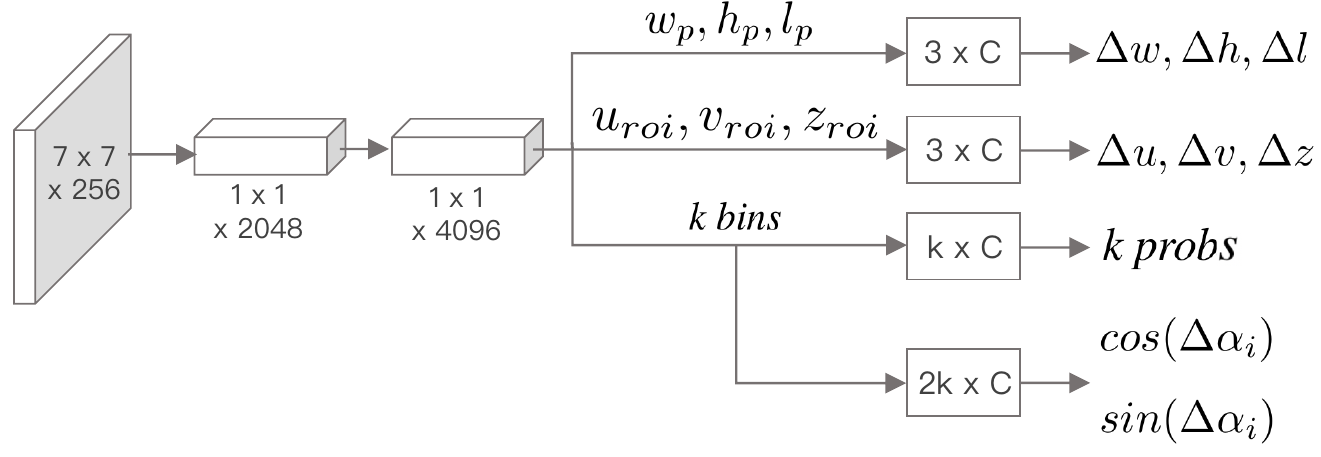}
		\caption{The data flow of the monocular 3D object detection. All above terms are defined in Sect.~\ref{sec:mono}.}
		\label{fig:notation}
	\end{figure}

	Our monocular 3D detector adopts a similar architecture with Faster R-CNN. We use identical region proposal network (RPN) for 2D RoIs generation. After RoI Align \cite{he2017mask}, we feed feature maps into the R-CNN head to extract high-level semantic information. Besides object classification and 2D box regression, we leverage additional fully-connected (\textit{fc}) layers to predict several spatial properties to recover the 3D bounding box. Directly regressing the 3D object pose is ill-posed from the cropped image RoI due to the loss of both location and size information. We thereby employ a \textit{residual} based regression to predict the 2D projection and depth of the 3D object center. Similar to 2D box prediction, we regress the normalized residual between the 2D projection of the 3D box center and the 2D RoI center:
	\begin{equation}
	\begin{aligned}
	\label{eq:center}
	\begin{array}{lr}
	\Delta u = (u_o - u_{\rm roi})/w_{\rm roi},\\
	\Delta v = (v_o - v_{\rm roi})/h_{\rm roi},
	\end{array}
	\end{aligned}  \rm with
	\begin{aligned}
	\begin{array}{lr}
	u_o = f_x \frac{x_o}{z_o} - u_p,\\
	v_o = f_y \frac{y_o}{z_o} - v_p,
	\end{array}
	\end{aligned}
	\end{equation}
	where $u_o, v_o$ denote the 2D projection coordinates of the 3D object center on the image, which is determined by the object location $[x_o,y_o,z_o]$, camera focal length $f_x, f_y$ and the principle point $u_0, v_0$ along the $u, v$ axis respectively. $u_{\rm roi}, v_{\rm roi}, w_{\rm roi}, h_{\rm roi}$ represent the center coordinates and the width and height of the 2D RoI. The residual of the object depth $z_o$ is defined as
	\begin{align}
	\label{eq:depth}
	\Delta z = \log(\tfrac{z_o}{z_{\rm roi}}),  \,z_{\rm roi} = f_y\tfrac{h}{h_{\rm roi}},
	\end{align}
	where $z_{\rm roi}$ can be considered as a coarse depth which is inferred from the perspective relation between the 3D object height $h$ and 2D RoI height $h_{roi}$. We define the dimension regression term as $\Delta d = \log \frac{({d}-{p_d})}{\sigma_d}, d \in  (w, h, l)$, where $p_d$ is a dimension prior and $\sigma_d$ for the corresponding standard deviation. As the global orientation $\theta$ is unobservable from the local image patch \cite{mousavian20173d, kundu20183d, li2019stereo}, we predict the observation angle $\alpha$ which can be uniquely determined by object appearance. 
	Inspired from \cite{mousavian20173d}, we employ \textit{MultiBin} based orientation estimation strategy. We classify the probability that the target angle lies in \textit{k} bins and regress the cosin and sine offset between the target angle and the bin's ray direction. In summary, we have $k$ classification terms for the \textit{bin} prediction, and $6+2k$ 3D box regression terms $[\Delta u, \Delta v, \Delta z, \Delta w, \Delta h, \Delta l, cos(\Delta \alpha_i), sin(\Delta \alpha_i)], i\in (1,$ $ \cdots,k)$ for each object category as illustrated in Fig.~\ref{fig:notation}.
	
	With these predicted variables, the 3D object location can be recovered using the inverse transforming as described in Eq.~\ref{eq:center},~\ref{eq:depth}. The global orientation $\theta$ is solved by decoupling the object location and the observation angle \cite{li2019stereo}. Unlike \cite{cvpr18xu} which requires depth input and \cite{qin2018monogrnet} employs an additional network for instance depth estimation, our monocular 3D object detector is light-weight and straightforward while achieves competitive performance on the KITTI~\cite{geiger2012we} benchmark. We use cross-entropy loss for object category and bin classification, smooth $L_1$ loss for all the regression terms. Note that we use overlapped orientation bins, so we apply the sigmoid function to the confidence outputs rather than the softmax. 
	
	\subsection{Stereo Photometric Alignment}
	\label{sec:stereo}
	The monocular image contains rich semantic information and therefore is suitable for high-level perception (e.g. foreground detection, classification). However, a single image is not able to provide accurate depth information for 3D localization. Considering this, we design our 3D object estimation framework as unified exploiting additional sensor data to refine the 3D localization.
	%Stereo imagery is a low-cost whereas effective solution for providing range information. 
	
\begin{figure}
		\center
		\vspace{0.2cm}
		\includegraphics[width=0.9\linewidth]{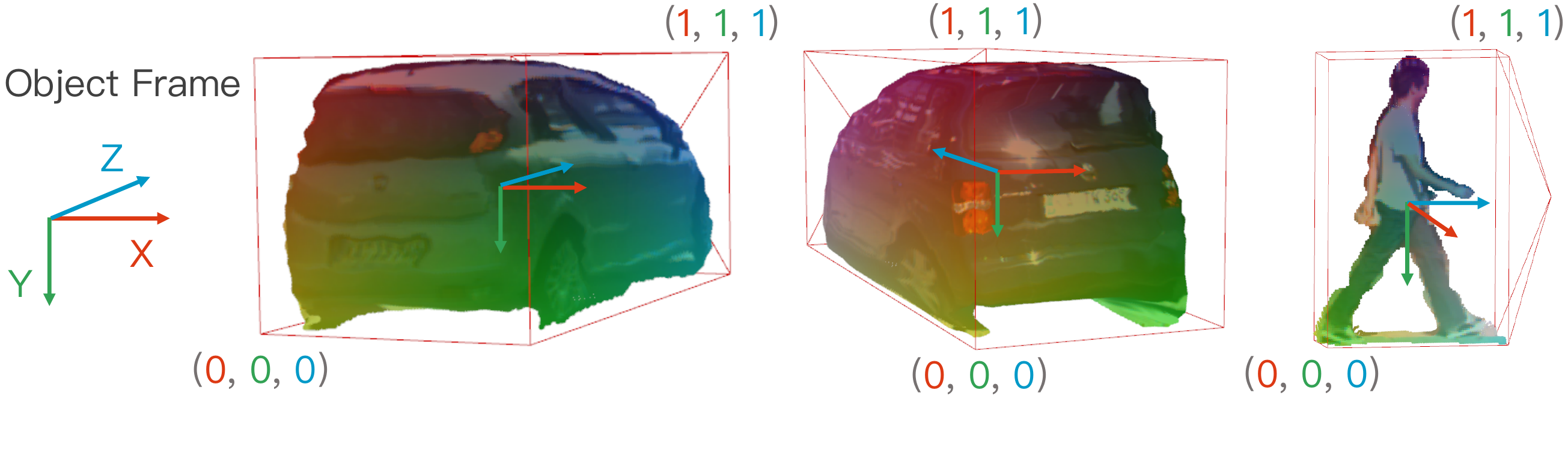}
		\caption{Illustration of the instance vector definition, where we normalize the physical coordinates to $[0,1]$ for consistency.}
		\label{fig:instance}
	\end{figure}

	If the stereo imagery is available, we expect to utilize it as a flexible and light-weight supplemental scheme for 3D localization enhancement instead of redeveloping a brand new stereo 3D detector. Preprocessing the stereo imagery as a dense disparity map is not only computational redundant for the only focus on the object area but also does not explicitly utilizing the object prior (e.g., object shape). Our previous work Stereo R-CNN \cite{li2019stereo} provides a heuristic for 3D bounding box refinement using raw stereo images, where the object depth is estimated according to the minimum warping cost of the object patch from the left image to the right image. However, treating the object as a regular cube unavoidably introduces shape error and limits the method to apply on general object categories (e.g., pedestrian, cyclist). To overcome this, we propose a simple approach to reconstruct the object local shape by predicting the element-wise instance vector, which represents each pixel's normalized 3D coordinates respecting to the object instance frame, which is illustrated in Fig.~\ref{fig:instance}. Similar representations are also used in \cite{behl2017bounding,wang2019normalized}. We normalize the physical coordinate to $[0,1]$ using the object dimension $\mathbf{d}_o$, which avoids the scale inconsistency issue among all pixels and multiple categories.
	
\begin{figure}
		\begin{center}
			\vspace{0.2cm}
			\includegraphics[width=0.7\columnwidth]{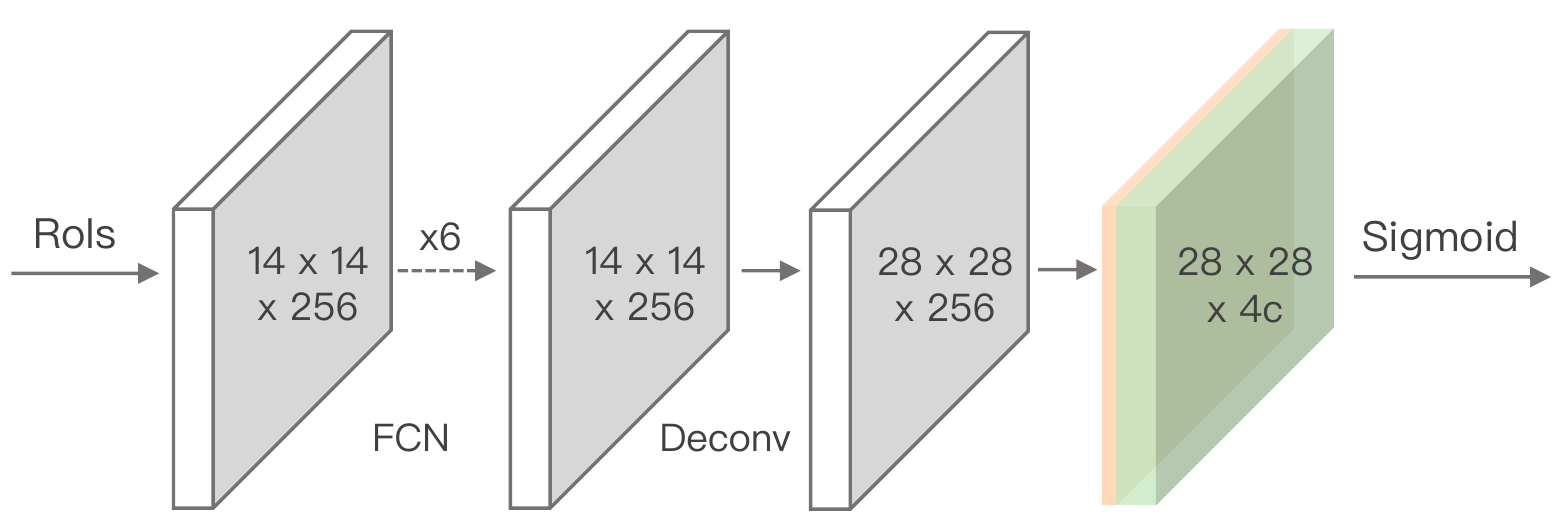}
		\end{center}
		%\vspace{-0.4cm}
		\setlength{\belowcaptionskip}{-0.2cm}
		\caption{RCNN head architecture for the pixel-wise instance vector predction and mask segmentation.}
		\label{fig:fcn}
	\end{figure}

	We adopt region-based FCN architecture proposed in \cite{he2017mask} for element-wise instance vector prediction and mask segmentation. As illustrated in Fig.~\ref{fig:fcn}, after RoI Align, 14 $\times$ 14 RoI feature maps are fed to six consecutive 256-d 3 $\times$ 3 convolution layers, each followed by a ReLU layer. We use a 2 $\times$ 2 deconvolution layer with stride 2 to upsample the output scale to $28 \times 28 \times 4\rm c$ which contains one segmentation channel and three instance vector channels corresponding to $\rm c$ categories. Due to we predict normalized instance vector, we apply element-wise \textit{sigmoid} to limit the output range between $[0,1]$.
	
	With the masked instance vector, we are able to recover the object local shape and perform the dense stereo alignment. Let $\mathbf{u}_i:=[u,v]^T$ represents the $i^{th}$ foreground pixel in the left image, which holds the instance vector $\mathbf{v}_i$. We 
	firstly recover its physical coordinate $^o\mathbf{p}_i$ in the object frame by scaling it with the regressed dimension $\mathbf{d}$ with element-wise multiplication $\otimes$:
	\begin{align}
	\label{eq:recover}
	^o\mathbf{p}_i &= \mathbf{v}_i\otimes\mathbf{d} - \tfrac{\mathbf{d}}{2}.
	\end{align}
	
	Gathering $^o\mathbf{p}_i$ for all foreground pixels forms the dense object shape, the stereo alignment error $\mathbf{E}_s$ can then be defined as the Sum of Squared Difference (SSD) of foreground pixels over the left and right image. 
	%Given the 2D prejection coordinate $\mathbf{u}_o:=[u,v]^{\rm T}$ of the object center, the dimension $\mathbf{d}_o$ and the rotation $\mathbf{R}(\theta)$ which are both predicted by the monocular regression (Sect. \ref{sec:mono}), the instance vector $\mathbf{v}_i$ for each forgrouund pixel, 
	Let
	\begin{align}
	\label{eq:stereo_error}
	\mathbf{E}_s &:= \sum^{n}_{i=0}  \left \| I_l(\mathbf{u}_i) - I_r(\pi(^c\hat{\mathbf{p}}_i - \mathbf{b})) \right \|,
	\end{align}
	where we use $\pi(\mathbf{p})$ to denote projecting a 3D point $\mathbf{p}$ to the image coordinate, and $\pi^{-1}(\mathbf{u},z)$ the corresponding inverse transformation, $z$ the depth of the pixel $\mathbf{u}$. 
	$^c\hat{\mathbf{p}}_i$ stands for the estimated pixel's 3D coordinate respecting to the left camera frame, given by
	\begin{equation}
	\begin{aligned}
	\label{eq:anchor}
	\begin{array}{lr}
	^c\hat{\mathbf{p}}_i = \hat{\mathbf{p}}_o + \mathbf{R}(\theta) ^o\mathbf{p}_i,
	\end{array}
	\end{aligned}  \rm with
	\begin{aligned}
	\begin{array}{lr}
	\hat{\mathbf{p}}_o = \pi^{-1}(\mathbf{u}_o,\hat{z}_o),
	\end{array}
	\end{aligned}
	\end{equation}
	where $\hat{\mathbf{p}}_o$ stands for the target 3D object location parameterized by the regressed projection $\mathbf{u}_o$ and the objective $\hat{z}_o$. 
	%To compensate the possible dimension ambiguity in monocular regression, we scale $^o\mathbf{p}_i$ using $s = {\hat{z}_o}/{{z}_o}$ according to the estimated depth $\hat{z}_o$ and regressed depth $z_o$. 
	Using above equations, each foreground pixel $^o\mathbf{p}_i$ is reprojected to the left camera frame according to the target object pose $[\hat{\mathbf{p}}_o, \mathbf{R}(\theta)]$, further to the right camera frame according to the stereo baseline $\mathbf{b} = [b,0,0]^T$. As a result, we are able to rectify the 3D object depth $\hat{z}_o$ by minimizing the photometric error $\mathbf{E}_o$ of a \textit{bundle} of pixel measurements.
	
	Comparing with directly regressing the object pose in 3D space, the instance vector representation has lower variance and is associated with distinct visual cues (e.g., rear lamps, wheels). It enables us to explicitly utilize the shape prior for multi-class objects, as well as exploit geometry information in raw stereo images to achieve superior accuracy as shown in Tab.\ref{table:stereo_SOTA}.
	
	\subsection{Point Cloud Alignment}
	\label{sec:lidar}
	Based on the unified geometric relations between sensor measurements with the 3D object center (Eq.~\ref{eq:anchor}), we can seamlessly extend the use of instance vector representation to the raw LiDAR point cloud. 
	%Similarly, We still treat the LiDAR as a flexible \textit{plug-in} sensor by leveraging our instance vector representation for the point cloud.
	To benefit from our existing monocular detection, we borrow a similar pipeline from F-PointNet \cite{qi2017frustum} for the point cloud processing. As overviewed in Fig.~\ref{fig:system}, we first collect the LiDAR points inside the 2D-lifted frustum based on 2D detections. A point-wise segmentation network is then used for masking the foreground points.
	% and use concatenated PointNets \cite{qi2017pointnet} for mask segmentation and point-wise instance vector regression.  
	After that, instead of regressing object center as proposed in \cite{qi2017frustum}, we use PointNet \cite{qi2017pointnet} to predict the identical instance vector representation as defined in Fig.~\ref{fig:instance}. Due to the geometric structure from the point cloud provides additional object size and orientation information, we keep the dimension and orientation regression branches in the original F-PointNet \cite{qi2017frustum}.
	\iffalse
	\begin{figure}
		%\setlength{\belowcaptionskip}{-0.2cm}
		\center
		\includegraphics[width=0.9\linewidth]{pointnet}
		\caption{PointNet architecture for the point-wise instance vector predction and mask segmentation.}
		\label{fig:pointnet}
	\end{figure}
	\fi
	We first resample the points in the frustum to a fix number $n$, as implemented in \cite{qi2017frustum}. A point-wise instance segmentation network is then employed to classify the foreground points. Using the predicted mask probability, we resample the foreground points to a fix number $m$. A light-weight T-Net \cite{qi2017frustum} is then applied to predict a coarse object center, which tries to align the centroid of foreground points closer to the object center to reduce variances. We refer readers to \cite{qi2017frustum} for more network details. With the aligned foreground points, we employ an instance vector regression branch which holds identical structure with the points segmentation network in parallel with the dimension and orientation regression branches. An element-wise \textit{sigmoid} function is applied to the instance vector to limit its range between $[0,1]$.
	
	Similarly with the stereo alignment, we use the point-wise instance vector $\mathbf{v}_i$ and the regressed dimension $\mathbf{d}$ to recover the spare local shape $^o\mathbf{p}_i$ using Eq.~\ref{eq:recover}. The 3D object can then be relocalized by aligning the local shape to the raw point cloud.
	We define the point cloud alignment error $\mathbf{E}_p$ as the sum of Euclidean distance between reprojected foreground points with the raw point cloud. Let
	\begin{equation}
	\begin{aligned}
	\label{eq:point_error}
	\mathbf{E}_p:= \sum^{m}_{i=0} \left \| ^c\mathbf{p}_i - {^c}\hat{\mathbf{p}}_i \right \|,\,
	\end{aligned}  \rm with
	\begin{aligned}
	\begin{array}{lr}
	^c\hat{\mathbf{p}}_i = \hat{\mathbf{p}}_o + \mathbf{R}(\theta) ^o\mathbf{p}_i,
	\end{array}
	\end{aligned}
	\end{equation}
	where $^c\mathbf{p}_i$ is the raw LiDAR point represented in the camera frame, ${^c}\hat{\mathbf{p}}_i$ stands for the estimated 3D point according to each point's local coordinate $^o\mathbf{p}_i$ and the object pose $[\hat{\mathbf{p}}_o, \mathbf{R}(\theta)]$. Note that we can recover the complete 3D location $\hat{\mathbf{p}}_o$ using the point cloud instead of only solving the center depth $\hat{z}_o$ in stereo alignment (refer Eq.~\ref{eq:anchor}). Since the orientation is regressed and treated as constant, Eq.~\ref{eq:point_error} can be solved linearly. By minimizing the distance between all estimated points ${^c}\hat{\mathbf{p}}_i$ and the raw point cloud ${^c}{\mathbf{p}}_i$, we recover the accurate 3D object location which best fits the object shape, which can also be viewed in Fig.~\ref{fig:cover}.
	
	\subsection{Multi-Sensor Fusion}
	\label{sec:fusion}
	Using our 3D object estimation framework, we formulate unified spatial relations between the 3D object center and sensor elements, which are data-agnostic and operate on the raw measurement domain. We can further jointly solve the Eq.~\ref{eq:point_error},~\ref{eq:stereo_error} using raw sensor uncertainty:
	\iffalse
	\begin{equation*}
	\small
	\begin{split}
	\renewcommand*{\arraystretch}{1}
	\label{eq:fusion}
	\mathbf{p}_o = \argmaxA_{\mathbf{p}_o}& \prod^{n}_{i=0}\prod^m_{j=0} p(I_l(\mathbf{u}_i)|\mathbf{p}_o, {^o}\mathbf{p}_i) p({^c}\mathbf{p}_j|\mathbf{p}_o, {^o}\mathbf{p}_j)\\
	=\argmaxA_{\mathbf{p}_o} &\sum^n_{i=0} \log p(I_l(\mathbf{u}_i)|\mathbf{p}_o, {^o}\mathbf{p}_i) + \sum^n_{i=0} \log p({^c}\mathbf{p}_j|\mathbf{p}_o, {^o}\mathbf{p}_i)\\
	=\argminA_{\mathbf{p}_o} &\sum^n_{i=0} \left \| I_l(\mathbf{u}_i) - I_r(\pi(^c\hat{\mathbf{p}}_i - \mathbf{b})) \right \|_{\sum_s} \\
	+ &\sum^{m}_{j=0} \left \| ^c\mathbf{p}_i - {^c}\hat{\mathbf{p}}_i \right \|_{\sum_p},
	\end{split}
	\normalsize
	\end{equation*}
	\fi
	\begin{equation}
	\small
	\begin{split}
	\renewcommand*{\arraystretch}{1}
	\label{eq:fusion}
	\mathbf{p}_o =\argminA_{\mathbf{p}_o} &\sum^n_{i=0} \left \| I_l(\mathbf{u}_i) - I_r(\pi(^c\hat{\mathbf{p}}_i - \mathbf{b})) \right \|_{\sum_s} + \\
	&\sum^{m}_{j=0} \left \| ^c\mathbf{p}_j - {^c}\hat{\mathbf{p}}_j \right \|_{\sum_p},
	\end{split}
	\normalsize
	\end{equation}
	where ${\scriptstyle{{\sum}_s}}, {\scriptstyle{\sum_p}}$ stands for the variance of image intensity and point cloud respectively. Weighting individual error term with the raw sensor uncertainty, Eq.~\ref{eq:fusion} enables us to flexibly fuse multiple sensors in a natural way to jointly refine the 3D object location $\hat{\mathbf{p}}_o$. Since our formulation uses unified geometric constraints from raw sensor measurements, it can be potentially applied to more general sensor configurations such as multiple stereo cameras and LiDARs.
	
	\section{Implementation Details}
	\label{sec:detail}
	\subsection{Loss.} Since there is no ground truth for the instance vector and mask in the KITTI object dataset \cite{geiger2012we}, we calculate the label with the aid of raw LiDAR point cloud and the ground-truth 3D bounding box. We first select points which lies within the 3D bounding box as foreground samples, the point-wise instance vector $\mathbf{v}_i$ is then generated by
	\begin{align}
	\label{eq:vector}
	\mathbf{v}_i = ({^o}\mathbf{p}_i + \tfrac{\mathbf{d}}{2}) \oslash\mathbf{d},\,\, {^o}\mathbf{p}_i = \mathbf{R}(\theta)^{-1}({^c}\mathbf{p}_i - \mathbf{p}_o)
	\end{align}
	where we transforming each LiDAR point ${^c}\mathbf{p}_i$ from the camera frame to the object frame ${^o}\mathbf{p}_i$ using the ground-truth object pose $[\mathbf{p}_o, \mathbf{R}(\theta)]$. $\oslash$ stands for the element-wise division for normalizing the physical coordinate ${^o}\mathbf{p}_i$ to $[0,1]$. With the point-wise label for the foreground points from Eq.~\ref{eq:vector}, we simply add an additional regression loss for the instance vector together with the other loss terms as defined in \cite{qi2017frustum}. 
	
	For the image, we label the sparse pixels which locate at the 2D projection of the foreground point cloud as the positive mask with the corresponding instance vector. Pixels which locate in the projection of the background points are labeled as the negative mask. 
	%Pixels without the corresponding 3D point cloud are ignored during training. 
	We define the multi-task loss for jointly training the monocular 3D detection, the pixel-wise mask and instance vector for the image:
	\begin{equation}
	\label{eq:loss}
	\renewcommand*{\arraystretch}{1.5}
	\begin{array}{lr}
	{L} = w_{rpn}L_{rpn} + w_{box}L_{box} + w_{orien}L_{orien} \\
	\,\,\,\,\,\,+\, w_{loc}L_{loc} + w_{dim}L_{dim},
	\end{array}
	\end{equation}
	where $L_{rpn}, L_{box}, L_{orien}$ denote losses for RPN, 2D detection, orientation respectively which contains both cross-entropy loss (objectness, multi-class, \textit{bin} classifications) and smooth $L_1$ loss. $L_{loc}, L_{dim}$ are smooth $L_1$ losses for the object location and dimension respectively. Each loss is weighted by their uncertainty as proposed in \cite{kendall2017multi}.

	\subsection{Training.} We use ResNet-101 \cite{he2016deep} with FPN \cite{lin2017feature} as the backbone network for extracting image feature. To incorporate spatial information for 3D object inference, we concatenate pixel-wise image $u, v$ coordinates as two additional channels with the raw RGB image. For initializing the network, we simply duplicate two channels in the first convolutional layer in the imagenet-pretrained weight.
	%The shorter side of the input image is resized to 600 pixels to handle small objects. 
	We augment the data by flipping the image, where the observation angle and the instance vector are mirrored correspondingly. For point cloud, we adopt the PointNet \cite{qi2017pointnet} and build on top of the baseline model in \cite{qi2017frustum} (denoted as v1 in the original paper). 
	%The sub-network of the instance vector prediction holds the same structure with the segmentation net, while takes 512 points as input for each instance. 
	
	The image and point cloud networks are trained separately since they share no feature map and gradient. We train the image network using SGD optimizer with 1 image and 512 RoIs per-batch for multi-class. The learning rate is set to 0.001 for the first 10 epochs, and reduced to 0.0001 for another 2 epochs. The PointNet is trained with exactly same hyper-parameters with the implementation in \cite{qi2017frustum}.
	\begin{table*}
		\begin{center}
			\vspace{0.2cm}
			\renewcommand{\arraystretch}{1.3}
			\resizebox{1.0\textwidth}{!}{%
				\begin{tabular}{l|ccc|ccc|ccc|ccc|ccc|ccc}
					\, &
					\multicolumn{9}{c|}{3D Bounding Box (AP$\rm _{3d}$)} & \multicolumn{9}{c}{Bird's Eye View (AP$\rm _{bv}$)}  \\
					\cline{2-19}
					\, &
					\multicolumn{3}{c|}{Car} & \multicolumn{3}{c|}{Pedestrian} & \multicolumn{3}{c|}{Cyclist} & \multicolumn{3}{c|}{Car} & \multicolumn{3}{c|}{Pedestrian} & \multicolumn{3}{c}{Cyclist} \\
					\cline{2-19}
					Method & \,Easy\, & Mode & \,Hard\, &  \,Easy\, & Mode & \,Hard\, & \,Easy\, & Mode & \,Hard\, &  \,Easy\, & Mode & \,Hard\, &  \,Easy\, & Mode & \,Hard\, & \,Easy\, & Mode & \,Hard \\
					\Xhline{1pt}  
					Mono3D \cite{chen2016monocular} &  2.53 & 2.31 & 2.31 & - & - & - & - & - & - &5.22 & 5.19 & 4.13 & - & - & - & - & - & -  \\
					Deep3DBox \cite{mousavian20173d} &  5.85 & 4.10 & 3.84 & - & - & - & - & - & - & 9.99 & 7.71 & 5.30 & - & - & - & - & - & -  \\
					Multi-Fusion \cite{cvpr18xu}  &  10.53 & 5.69 & 5.39 & - & - & - & - & - & - &  \textbf{22.03} & 13.63 & 11.60  & - & - & - & - & - & -  \\
					MonoGRNet \cite{qin2018monogrnet} &  13.88 & 10.19 & 7.62 & - & - & - & - & - & -& - &- & - & - & - & - & - & - & -  \\
					%ROI-10D \cite{} & 14.50 & 9.91 & 8.73 & - & - & - & - & - & - &  9.61 & 6.63 & 6.29 & - & - & - & - & - & - \\
					MonoPSR \cite{ku2019monocular} &  12.75 & 11.48 & 8.59 & \textbf{10.64} & 8.18 & 7.18& 10.88 & 9.93 & 9.93 & 20.63 & \textbf{18.67} & 14.45  & \textbf{11.68} & 10.05 & 8.14 & 11.18 & 10.18 & 10.03  \\
					\hline
					%Ours* (Mono) &\textbf{18.38} & \textbf{14.20} & \textbf{13.09} & 10.43 & \textbf{10.12} & 9.30 & 4.62 & 3.26 & 2.84&  \textbf{23.05} & 17.25 & 15.42  &10.69 & \textbf{10.38} & 10.04 & 13.07 &11.01 & 10.77  \\
					Ours (Mono) &  \textbf{16.02} & \textbf{13.79} & \textbf{12.03} & 10.34 & \textbf{9.88} & \textbf{9.91} & \textbf{13.16} & \textbf{11.03} & \textbf{11.02} &  20.10 & 16.04 & \textbf{15.61}  &10.75 & \textbf{10.37} & \textbf{10.06} & \textbf{13.51} & \textbf{11.21} & \textbf{11.25}  \\
					
				\end{tabular}
			}		
		\end{center}
		\caption{Comparing the monocular 3D localization and detection performance with state-of-the-arts, evaluated on KITTI \textit{val} set.}
		\label{table:mono_SOTA}
	\end{table*}
	
	\begin{table*}
		\begin{center}
			\renewcommand{\arraystretch}{1.3}
			\resizebox{1.0\textwidth}{!}{%
				\begin{tabular}{l|ccc|ccc|ccc|ccc|ccc|ccc}
					\, &
					\multicolumn{9}{c|}{3D Bounding Box (AP$\rm _{3d}$)} & \multicolumn{9}{c} {Bird's Eye View (AP$\rm _{bv}$)} \\
					\cline{2-19}
					\, &
					\multicolumn{3}{c|}{Car} & \multicolumn{3}{c|}{Pedestrian} & \multicolumn{3}{c|}{Cyclist} & \multicolumn{3}{c|}{Car} & \multicolumn{3}{c|}{Pedestrian} & \multicolumn{3}{c}{Cyclist} \\
					\cline{2-19}
					Method & \,Easy\, & Mode & \,Hard\, &  \,Easy\, & Mode & \,Hard\, & \,Easy\, & Mode & \,Hard\, &  \,Easy\, & Mode & \,Hard\, &  \,Easy\, & Mode & \,Hard\, & \,Easy\, & Mode & \,Hard \\
					\Xhline{1pt}  
					3DOP \cite{3dopJournal} &  6.55 & 5.07 & 4.10 & - & - & - & - & - & - & 12.63 & 9.49 & 7.59 & - & - & - & - & - & -  \\
					Multi-Fusion* \cite{cvpr18xu} &  - & 9.80 & - & - & - & - & - & - & -  & - & 19.54 & - & - & - & - & - & - & - \\
					TLNet \cite{qin2019tlnet} &  18.15 & 14.26 & 13.72 & - & - & - & - & - & - & 29.22 & 21.88 & 18.33  & - & - & - & - & - & -  \\
					Stereo R-CNN \cite{li2019stereo} &  54.11 & {36.69} & 31.07 & - & - & - & - & - & - & 68.50 & {48.30} & 41.47 & - & - & - & - & - & -  \\
					Pseudo-3D \cite{wang2018pseudo} & 56.70 & 37.90 & 34.30 & 23.50 & 19.40 & 15.30 & 28.50 & 19.30 & 18.20 & {74.00} & \textbf{54.70} & \textbf{47.30} & 32.50 & 27.10 & 23.10 & 35.40 & 23.70 &  \textbf{22.00}  \\
					\hline
					%Ours* (Stereo) &  \textbf{58.40} &   \textbf{37.59} & 31.64 & 27.44 & 23.59 & 21.01 &  \textbf{30.75} &  \textbf{20.47} &  \textbf{19.12} &  68.58 & 48.14 & 40.49 & 33.21 & 28.98 & 24.39  & 33.81 & 22.47 & 21.36  \\
					%Ours (Stereo) & 54.51 & 36.53 & \textbf{32.87} & \textbf{30.87} & \textbf{26.01} & \textbf{22.80} & 30.56 & 20.01 & 18.79 &  66.82 & 46.40 & 38.54  &\textbf{36.91} & \textbf{32.28} & \textbf{27.68}&\textbf{36.56} & \textbf{24.11} & \textbf{22.72}  \\
					Ours (Stereo) & \textbf{61.96} &  \textbf{42.03} &  \textbf{34.57} &  \textbf{33.65} &  \textbf{28.60} &  \textbf{25.47} &  \textbf{31.60} &  \textbf{20.58} &  \textbf{19.12} &  \textbf{75.13} &  54.19 &  45.12  &  \textbf{41.62} &  \textbf{35.95} &  \textbf{29.74} &  \textbf{36.30} &  \textbf{24.02} & 21.85  \\
				\end{tabular}
			}		
		\end{center}
		%\vspace{-0.4cm}
		\caption{Stereo 3D detection comparison using bird's eye view AP$_{\rm bv}$ and 3D boxes AP$_{\rm 3d}$, evaluated on the KITTI \textit{val} set. Multi-Fusion* \cite{cvpr18xu} indicates its improved version using the stereo depth as input.}
		\label{table:stereo_SOTA}
	\end{table*}
	
	\section{Experiments}
	We evaluate our 3D object estimation framework on the KITTI object benchmark \cite{geiger2012we}, which contains 7481 training samples and 7518 testing samples. 
	Since the ground truth of test set if unavailable, we divide the training samples into \textit{train} and \textit{val} split following \cite{3dopJournal}. Considering the KITTI test server limits 3 submissions per month, we evaluate our monocular, stereo 3D performance on the \textit{val} set as previous image-based methods do, and submit our LiDAR-based results to the test server. We compare the 3D detection (3D bounding box AP$\rm _{3d}$) and 3D localization (bird's eye view AP$\rm _{bv}$) performance for car, pedestrian, cyclist with all state-of-the-are methods. We set IoU threshold to 0.7 for cars and 0.5 for pedestrians and cyclists for all experiments.
	
	\subsection{Monocular Evaluation.}We first evaluate our fundamental monocular 3D detection performance. As reported in Table.~\ref{table:mono_SOTA}, we compare with the state-of-the-art monocular methods using Average Precision for 3D box AP$\rm _{3d}$ and bird's eye view AP$\rm _{bv}$ on three categories, where Multi-Fusion \cite{cvpr18xu} requires dense depth image as input. MonoGRNet \cite{qin2018monogrnet} and MonoPSR \cite{ku2019monocular} exploit sophisticated feature fusion strategy to encode multi-level information, while our simple-designed 3D detector is general for multiple categories and require few additional computations comparing to typical 2-stage 2D detectors. We show the overall best performance in 3D localization task and 3D detection task among car, pedestrian, and cyclist classes.
	
	\subsection{Stereo Evaluation.}Benefited from our unified 3D object estimation framework, the stereo data can be seamlessly introduced into our system to refine the 3D localization accuracy. With the aid of our pixel-level stereo alignment module (Sect.~\ref{sec:stereo}), the 3D detection performance can be improved significantly.
	%Results are reported in Table.~\ref{table:stereo_SOTA}, where Pseudo-3D \cite{wang2018pseudo} and Stereo R-CNN \cite{li2019stereo} are the most recent state-of-the-art stereo-based methods. Pseudo-3D \cite{wang2018pseudo} utilizes LiDAR-based methods \cite{ku2017joint,qi2017frustum} to localize the 3D object by converting the stereo-generated disparity to the point cloud. 
	The 3D bounding box AP$\rm _{3d}$ and bird's eye view AP$\rm _{3d}$ comparison results are reported in Table.~\ref{table:stereo_SOTA}.
	We list the Pseudo-3D \cite{wang2018pseudo} result as reported in their conference submission version, which utilizes LiDAR-based methods \cite{ku2017joint,qi2017frustum} to localize the 3D object by converting the stereo-generated disparity to point cloud representation. Our previous work Stereo R-CNN \cite{li2019stereo} exploits sparse 2D bounding box and keypoints constraints to solve initial 3D box and refine object depth by dense photometric alignment. However, the box-shape approximation introduces modeling error and limits its usage on the car only. 
	Thanks to our instance-vector based shape-aware alignment module, our method is naturally general for multiple categories objects. As Table.~\ref{table:stereo_SOTA} reported, we show better 3D detection and localization performance compared with previous state-of-the-art stereo methods.
	
	Note that the easier regime the object belongs, we obtain more remarkable improvements compared with other methods. This phenomenon meets our expectation since it is easier to learn a better shape for nearby objects because of more observations and denser instance vector labels.
	We train the car category separately for a fair comparison, all other experiments are trained jointly.
	
	\begin{figure*}
		\setlength{\belowcaptionskip}{-0.2cm} 
		\begin{center}
			\vspace{0.3cm}
			\includegraphics[width=2\columnwidth]{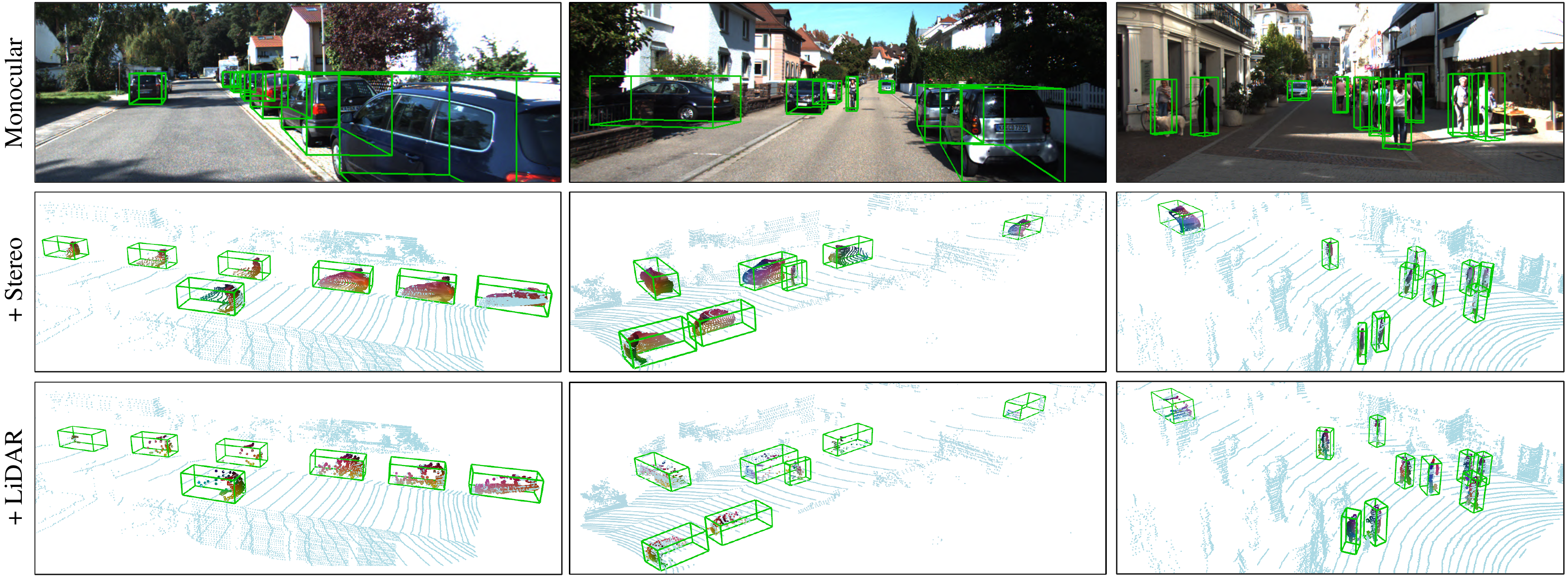}
		\end{center}
		%\vspace{-0.4cm}
		\caption{Qualitative examples (best view with zoom in). From top to bottom: 3D detection results using monocular, stereo images, and LiDAR point cloud, where we additionally visualize the reconstructed object shape using our instance vector representation.}
		%Enlarged details show that the object structure is well aligned with the raw point cloud.
		\label{fig:quali}
	\end{figure*}
	
	\subsection{LiDAR Evaluation.} To demonstrate our framework is general for multiple sensors, we extend it to the use of LiDAR point cloud. We first conduct comparisons with the baseline model F-PointNet (v1) \cite{qi2017frustum} which regresses the 3D object location while we solve the 3D object location by minimizing the point cloud alignment error (see Eq.~\ref{eq:point_error}). 
	As shown in Table.~\ref{tab:lidar_com}, when using the same 2D RoIs input, regressed orientation and dimension, our point cloud alignment method shows overall better performance on the 3D detection and localization tasks, which evidences again our instance vector based estimation is not only general for multiple sensors but also effective on the accuracy.
	
	We further submit our result to the KITTI \cite{geiger2012we} test server. As detailed in Table.~\ref{table:lidar_SOTA}, we show consistent improvements comparing with the baseline F-PointNet (v1) \cite{qi2017frustum}.
	%As detailed in Table.~\ref{table:lidar_SOTA}, we list AP$_{\rm bv}$ and AP$_{\rm 3d}$ of recent state-of-the-art LiDAR-based methods according to the reports in their papers.
	We also list other state-of-the-art LiDAR-based methods for reference. Since our system utilizes LiDAR in the second stage only, it unavoidably underperforms the full LiDAR-based methods.
	Note that this letter focuses on a unified geometric 3D object estimation framework for multiple sensors and multiple categories rather than the single modal accuracy. Fully exploiting the LiDAR modality in RPN stage (\cite{shi2018pointrcnn}) or employing a more sophisticated network (e.g., PointNet++ \cite{qi2017pointnet++}) may further improve the 3D APs while outside the scope of this work.
	
	\subsection{Comparing Different Sensor Combinations.} As described Sect.~\ref{sec:fusion}, our 3D estimation framework unified use different input and can perform sensor fusion in a natural way. We compare the 3D localization performance for different sensor combinations in Table.~\ref{tab:fusion}. Same RoIs (provided in \cite{qi2017frustum}) are adopted for all sensor combination to achieve clear comparison. Note that the effective range of the stereo is much smaller than the LiDAR in the KITTI setting, thus the fusion of LiDAR and stereo brings trivial improvements. We expect our uncertainty-aware estimation yields more promising gains when dealing with more general and balanced multi-sensor settings in modern self-driving cars (e.g., multiple stereos with different baseline, multiple overlapped LiDARs).
	
	\begin{table}
		\setlength{\belowcaptionskip}{-0.2cm} 
		\centering
		\renewcommand{\arraystretch}{1.3}
		\resizebox{0.45\textwidth}{!}{%
			\begin{tabular}{l|ccc|ccc}
				\, & \multicolumn{3}{c|}{Baseline \cite{qi2017frustum}} & \multicolumn{3}{c}{Ours (LiDAR)}\\
				\cline{2-7}
				Benchmark & \,Easy\, & Mode & \,Hard\, &  \,Easy\, & Mode & \,Hard\, \\
				\Xhline{1pt} 
				Car (AP$\rm _{bv}$)  & 87.90 & \textbf{82.45} & \textbf{74.31} & \textbf{88.29} & 82.05 & 73.59 \, \\
				Pedestrian (AP$\rm _{bv}$)  & 69.69 & 60.54 & 53.16 & \textbf{72.22} & \textbf{62.84} & \textbf{57.75} \, \\
				Cyclist (AP$\rm _{bv}$)  & 81.00 & 58.61 & 54.55 & \textbf{82.48} & \textbf{62.68} & \textbf{58.30} \, \\
				\hline
				Car (AP$\rm _{3d}$)  & 82.85 & 69.36 & 62.55 & \textbf{84.47} & \textbf{71.50} & \textbf{63.60} \, \\
				Pedestrian (AP$\rm _{3d}$)  & 65.15 & 55.41 & 48.52 & \textbf{67.51} & \textbf{57.90} & \textbf{50.92} \, \\
				Cyclist (AP$\rm _{3d}$)  & 76.62 & 54.29 & 50.35 & \textbf{78.18} & \textbf{57.93} & \textbf{54.36} \,
		\end{tabular}}
		\caption{Comparing our results with the baseline F-PointNet (v1) \cite{qi2017frustum}, where exactly same 2D boxes, orientation, and dimension are used for fair comparison, evaluated on the KITTI \textit{val} set.}
		\label{tab:lidar_com}
	\end{table}
	
	\begin{table}
		\centering
		\renewcommand{\arraystretch}{1.3}
		\resizebox{0.45\textwidth}{!}{%
			\begin{tabular}{l|ccc|ccc}
				\, & \multicolumn{3}{c|}{3D Box (AP$\rm _{3d}$)} & \multicolumn{3}{c}{Bird's View (AP$\rm _{bv}$)}\\
				\cline{2-7}
				Category & \,Car\, & Ped & \,Cyc\, &  \,Car\, & Ped & \,Cyc\, \\
				\Xhline{1pt} 
				VoxelNet \cite{zhou2017voxelnet} & 65.11 & 33.69 & 48.36 & 79.26 & 40.74 & 54.76 \, \\
				AVOD \cite{ku2017joint} & 71.88 & 42.81 & 52.18 & 83.79 & 51.05 & 57.48 \, \\
				F-PointNet (v2) \cite{qi2017frustum} & 70.39 & 44.89 & 56.77 & 84.00 & 50.22 & 61.96 \, \\
				PointPillar \cite{lang2018pointpillars} & 74.99 & 43.53 & 59.07 & 86.10 & 50.23 & 62.25 \, \\
				PointRCNN \cite{shi2018pointrcnn} & 75.42 & 41.78 & 59.60 & 86.04 & 47.53 & 66.77 \, \\
				\hline
				Baseline \cite{qi2017frustum} & 64.70 & 41.55 &53.50 & \textbf{77.09} & 47.56 &59.87 \\
				Ours (LiDAR) & \textbf{65.33} & \textbf{42.87} & \textbf{59.40} & 76.65 & \textbf{47.78} & \textbf{63.72}\\
		\end{tabular}}
		\caption{LiDAR 3D detection comparison using bird's eye view AP$_{\rm bv}$ and 3D boxes AP$_{\rm 3d}$, reported for three categories moderate objects on the KITTI \textit{test} set.}
		\label{table:lidar_SOTA}
	\end{table}
	
	\begin{table}
		\setlength{\belowcaptionskip}{-0.3cm} 
		\centering
		\renewcommand{\arraystretch}{1.3}
		\resizebox{0.37\textwidth}{!}{%
			\begin{tabular}{l|ccc}
				Input Data & \,Easy\, & Mode & \,Hard\,  \\
				\Xhline{1pt} 
				Monocular  & 23.05 & 17.25 & 15.42  \\
				Monocular + Stereo  & 74.88 & 52.40 & 44.23 \\
				Monocular + LiDAR  & 88.29 & 82.06 & 73.59  \\
				Monocular + Stereo + LiDAR  & \textbf{88.37} & \textbf{82.07} & \textbf{73.60}  \\
		\end{tabular}}
		\vspace{0.2cm}
		\caption{3D localization performance (AP$_{\rm bv}$) of using different data, evaluated on the KITTI \textit{val} set. We adopt the same 2D RoI input from \cite{qi2017frustum} among all combinations for clear comparisons, the APs for the monocular and stereo thereby show slightly different performance as we reported in Table.~\ref{table:mono_SOTA},~\ref{table:stereo_SOTA} where we use our 2D detector.}
		\label{tab:fusion}
	\end{table}

	\subsection{Qualitative Results.}We visualize some qualitative results in Fig.~\ref{fig:quali}. Considering the 3D object center is invisible from raw sensor data (only the surface can be directly observed), we formulate rigorous geometric constraints between observed sensor elements with the 3D object center. Therefore, our method encourages the alignment accuracy of the nearest object surface, which is more important for obstacle avoidance in autonomous driving scenarios.
	
	\section{Conclusion and Future Work}
	In this latter, we aim to detect the object in the image and improve the 3D localization accuracy using multi-sensor refinement in autonomous scenarios. Using our instance vector representation, we formulate the unified spatial relations between 3D object center and raw sensor measurements. Benefited from this, we are able to exploit the individual characteristics of the different sensor while constrain the 3D object location in a unified way, which further enables us to naturally fuse multiple raw sensor data. On the 3D detection and 3D localization tasks, we outperform previous state-of-the-art monocular, stereo methods and also demonstrate competitive performance compared with the baseline LiDAR-based method.
	
	Although our current system uses the monocular image for object proposal, our framework can seamlessly adopt 3D proposals from the point cloud to take the advantage of a high recall rate. which indicates the possible future direction of fully exploiting multiple sensors for more powerful object proposals.
	%One weakness is that our framework is only unified for 3D object localization with different sensors while still relies on the monocular image for region proposal, which indicates the possible future direction. We hope similar designing can be exploited for the object proposal,
	
	%-------------------------------------------------------------------------
	{
		% \small
		% \bibliographystyle{ieee}
		\bibliographystyle{unsrt}
		\bibliography{egbib}
	}
	
\end{document}